\documentclass[10pt,conference,a4paper]{IEEEtran}

\usepackage{times,amsmath,epsfig}
\usepackage{epstopdf}
\usepackage{color}
\usepackage{amssymb}
\usepackage{amsmath}
\usepackage{makecell,multirow,diagbox}
\usepackage{graphicx}
\usepackage[misc]{ifsym}
\title{Deep Hashing with Triplet Quantization Loss}
\author{%
{Yuefu Zhou, Shanshan Huang, Ya Zhang\textsuperscript{\Letter}, Yanfeng Wang\textsuperscript{\Letter}} 
{}
\vspace{1.6mm}\\
\fontsize{10}{10}\selectfont\itshape

\fontsize{10}{10}\selectfont\rmfamily\itshape
Cooperative Medianet Innovation Center, Shanghai Jiao Tong University, Shanghai, China\\ 
\fontsize{9}{9}\selectfont\ttfamily\upshape
\{remicongee,  
huangss,  
ya\_zhang,  
wangyanfeng\}@sjtu.edu.cn 

}
\begin{document}
\maketitle

\begin{figure}[b]
\parbox{\hsize}{\em
IEEE VCIP'17, Dec. 10 - Dec. 13, 2017, St. Petersburg, US

978-1-5386-0462-5/17/\$31.00 \ \copyright 2017 IEEE.
}\end{figure}

\begin{abstract}
With the explosive growth of image databases, deep hashing, which learns compact binary descriptors for images, has become critical for fast image retrieval. 
Many existing deep hashing methods leverage {\em quantization loss}, defined as distance between the features before and after quantization, to reduce the error from binarizing features. While minimizing the quantization loss guarantees that quantization has minimal effect on retrieval accuracy, it unfortunately significantly reduces the expressiveness of features even before the quantization. In this paper, we show that the above definition of quantization loss is too restricted and in fact not necessary for maintaining high retrieval accuracy. We therefore propose a new form of quantization loss measured in triplets. The core idea of the triplet quantization loss is to learn discriminative real-valued descriptors which lead to minimal loss on retrieval accuracy after quantization.
Extensive experiments on two widely used benchmark data sets of different scales, CIFAR-10 and In-shop, demonstrate that the proposed method outperforms the state-of-the-art deep hashing methods. Moreover, we show that the compact binary descriptors obtained with triplet quantization loss lead to very small performance drop after quantization.
\\[1\baselineskip]
\end{abstract}


\section{Introduction}

The last decade has witnessed an explosive growth in multimedia content, especially videos and images. As a result, hashing, which generates compact representative descriptors for fast image retrieval, has received much attention in computer vision. 
Locality-sensitive Hashing (LSH), which generates hash functions by random projection, has been widely adopted for hand-craft features. In recent years, it has become increasingly popular to learn hashing functions directly from training data. Iterative Quantization (ITQ) learns an orthogonal rotation on the transformed data to reduce the quantization distortion \cite{gong2013iterative}. Miminal Loss Hashing (MLH) learns hash functions by optimizing a hinge-like loss \cite{norouzi2011minimal}.  Binary Reconstructive Embedding (BRE) \cite{kulis2009learning} and Kernel-Based Supervised Hashing (KSH) \cite{liu2012supervised} adopt Kernel methods for hashing learning. With the advancement of deep learning techniques for image retrieval, more recent studies focus on deep hashing. Deep Neural Network Hashing (DNNH) learns to directly output binary codes \cite{lai2015simultaneous}. Convolutional Neural Network Hashing (CNNH) learns binary-like descriptors with similarity matrix \cite{xia2014supervised}. 

More recent deep hashing methods usually append a hashing layer to the typical Convolutional Neural Network (CNN) to learn low-dimensional descriptors and then generates compact binary codes via quantization. Triplet loss \cite{schroff2015facenet} is often adopted by these methods to generate discriminative real-valued descriptors. However, the subsequent quantization decreases the discriminability of features and results in a significant drop in retrieval accuracy. To reduce the error of binarizing features, many methods like Deep Supervised Hashing (DSH) \cite{liu2016deep} introduce {\em quantization loss} as the distance between real-valued features and the corresponding binary codes. Minimizing the quantization loss forces the CNN model to directly learn binary-like descriptors and hence reduces the performance drop due to subsequent quantization. Unfortunately, while it narrows the gap between real-valued features and the corresponding binary codes, it actually results in less discriminative real-valued features. In fact, the sufficient and necessary condition for minimizing the performance gap between features before and after binarizing is that the distance of similar images remains smaller than that of dissimilar images after quantization. 

In this work, we propose a supervised deep hashing method to address the above problem. In particular, we design a new quantization loss measured in triplets to implement the sufficient and necessary condition for hashing learning. The main idea of the triplet quantization loss is to modify the distribution of real-valued features according to the threshold for quantization. We first train a CNN model with the triplet loss and then fine-tune the model with triplet quantization loss so that the real-valued descriptors are not only discriminative but also suitable for quantization, i.e. the real-valued features learned are expected to lead to minimal loss on retrieval accuracy after quantization. To learn descriptors fitting given quantization function, we propose a method based on statics for setting parameters. We validate the effectiveness of the proposed method with two widely used benchmark data sets of different scales, \textbf{CIFAR-10} and \textbf{In-shop}. The experimental results have demonstrated that the proposed approach outperforms the state-of-the-art hashing methods. Moreover, we show that the compact binary descriptors obtained with triplet quantization loss lead to very small performance drop after quantization.

\section{Proposed Approach}

We employ the architecture proposed by Krizhevsky \emph{et al.} \cite{krizhevsky2012imagenet} for CNN model and append a fully-connected latent layer $F$ with \emph{N} neurons activated by sigmoid function for hashing. The learning strategy consists of two steps. The first step is for learning a low-dimension real-valued features. Similar to the existing methods, we adopt CNN model with the triplet loss for the first step. The second step is to optimize the features for subsequent quantization. We design a new quantization loss, triplet quantization loss, to fine-tune the CNN model. 
The details of our method are described as follows.

\subsection{Triplet Quantization Loss}

Similar to CNN based on triplet loss, for each anchor image $a$, we randomly choose one similar image $p$ and a dissimilar image $n$. In both steps, given the $i$-th triplet $(a_i,p_i,n_i)$, let $(f^{a_i},\,f^{p_i},\,f^{n_i})$ denote the output of the latent layer $F$.

To obtain expressive descriptors, we firstly train our model with the triplet loss 
\begin{equation}\label{L_triplet}
  L_{t}=\frac{1}{2bs}\sum_{i=1}^{bs}{\max{(\alpha+\left\|f^{a_i}-f^{p_i}\right\|^2-\left\|f^{a_i}-f^{n_i}\right\|^2,\,0)}},
\end{equation}
where $bs$ is the batch size, $\alpha$ is a parameter for margin.
For quantization, we define the binary code $b$ as 1 if the feature $f$ exceeds 0.5 and as 0 if not. 

Although the triplet loss is effective in learning real-valued descriptors with high discriminability, its discriminability is significantly lowered after quantization.
To make real-valued descriptors suitable for quantization, we propose a triplet quantization loss, a variant of the triplet loss. First of all, there is no need to force the feature values to be either 0 or 1, which dramatically decreases the expressiveness of the real-valued descriptors. In fact, the triplet quantization loss only requires that for similar images,  their feature values are both greater or smaller than the quantization threshold $t$ (0.5 in our case) in each dimension. The loss for similar pairs is formulated as:
\begin{equation}\label{L_sim}
  L_{s}=\frac{1}{bs}\sum_{i=1}^{bs}\sum_{j=1}^{N}\max{(\alpha_{s}-(f_{j}^{a_i}-0.5)(f_{j}^{p_i}-0.5),\,0)},
\end{equation}
where $\alpha_{s}$ is a parameter for slack.
Because it is difficult to find in which dimensions anchor and negative have their feature values on different ``sides" of 0.5, Euclidean distance is used in formulating the loss for dissimilar pairs:
\begin{equation}\label{L_dis}
  L_{d}=\frac{1}{2bs}\sum_{i=1}^{bs}\max{(\alpha_{d}-\sum_{j=1}^N \min({\left|f_{j}^{a_i}-f_{j}^{n_i}\right|^2,\,\delta}),\,0)},
\end{equation}
where $\alpha_{d}$ and $\delta$ are parameters for slack.

Combining Eq. \ref{L_sim} and Eq. \ref{L_dis}, the triplet quantization loss is formulated as
\begin{equation}\label{L_TQN}
    L_{TQN}=\beta\,L_{s}\,+\,\gamma\,L_{d}
\end{equation}
where $\beta$ and $\gamma$ are two parameters to balance different losses.

In particular, we compute the derivative of the loss $L_{TQN}$ with respect to features as follows:
\begin{equation}
\begin{split}
    \frac{\partial L_{TQN}}{f_j^{a_i}} = &\,\frac{\beta}{bs}\left(0.5 - f_j^{p_i}\right)\cdot1_{>0}(L_{s, i}) \\
    & + \frac{\gamma}{bs}\left(f_j^{a_i}-f_j^{n_i}\right)\cdot 1_{>0}(L_{d, i})\cdot 1_{>0}\left(\tau_{i, j}\right),
\end{split}
\end{equation}
\begin{equation}
    \frac{\partial L_{TQN}}{f_j^{p_i}} = \frac{\beta}{bs}\left(0.5 - f_j^{a_i}\right)\cdot 1_{>0}\left(L_{s,i}\right),
\end{equation}
\begin{equation}
    \frac{\partial L_{TQN}}{f_j^{n_i}} = \frac{\gamma}{bs}\left(f_j^{a_i}-f_j^{n_i}\right)\cdot 1_{>0}(L_{d, i})\cdot 1_{>0}\left(\tau_{i, j}\right),
\end{equation}
where $1_{>0}(x)$ outputs 1 if $x>0$ and 0 otherwise, $L_{s, i}$ (respectively $L_{d, i}$) denotes $i$-th item of $L_s$ (respectively $L_d$) and $\tau_{i, j}$ denotes $\delta-\left|f_{j}^{a_i}-f_{j}^{n_i}\right|^2$.

\subsection{Parameter Analysis}

In this section, we analyze the parameters $\alpha_{s}$ and $\alpha_{d}$ defined in our method and seek for reasonable settings for them.

\subsubsection{$\alpha_{s}$}
\

Suppose in each dimension, it is sufficient that anchor and positive's real-valued
features are at least $\Delta$ far from 0.5, which means, for the $i$-th similar pair
and in any dimension $j$, we have
\begin{equation}
\begin{split}
  \min&{(\left|f_{j}^{a_i}-0.5\right|,\,\left|f_{j}^{p_i}-0.5\right|)}>\Delta \\
  s.t&.\,(f_{j}^{a_i}-0.5)(f_{j}^{p_i}-0.5)>0 \\
  &\;\;\Delta\in \left[0,\,0.5\right].
\end{split}
\end{equation}
Given $L_{s}$ defined in Eq. \ref{L_sim}, it is reasonable to set
\begin{equation}
  \alpha_{s}=\Delta^2\;\;\;\;s.t.\,\Delta\in \left[0,\,0.5\right].
\end{equation}

\subsubsection{$\alpha_{d}$}

\

Suppose there are $C$ classes of images in our dataset. In order to
cover all these classes by binary codes, it needs at least $M=\left\lceil \log_2C\right\rceil$
bits.
Let $S_N$ be the space
of $N$-bit codes and $S_M$ be the space of $M$-bit codes.
Suppose we can ignore the approximation brought by $\left\lceil .\right\rceil$
and construct a surjective function
\begin{equation}\label{surjection}
\begin{split}
  \phi:\,S_N&\to S_M \\
  b&\mapsto B.
\end{split}
\end{equation}
On average, $N/M$ bits of $b$ contribute 1 bit in $B$.

Now suppose that each bit of $B$ follows Bernoulli distribution $\mathcal{B}\left(\frac{1}{2}\right)$
and independent of others. In $S_M$, for the codes of an anchor image $B^a$ and a negative image $B^n$, their expected hamming distance $d_H(B^a,B^n)$ can be calculated as follows.
\begin{equation}
  \mathbb{E}\left[d_H(B^a,B^n)\right]=\sum_{l=1}^Ml\binom{M}{l}p^l\left(1-p\right)^{M-l}=pM,
\end{equation}
where $p=\frac{2^{M-1}}{2^M-1}$.

Note that $\mathbb{E}\left[d_H(B^a,B^n)\right]$ is independent of $(a,n)$ and $M$ is usually big enough so that $p\thickapprox \frac{2^{M-1}}{2^M}=\frac{1}{2}$.
Hence, any pair of different codes in $S_M$ has $\frac{M}{2}$ different bits on average.

Considering the surjective projection $\phi$ defined in Eq. \ref{surjection}, it is safe to deduce
that 
\begin{equation}
  \forall (b^a,\,b^n)\in S_N^2,\,\mathbb{E}\left[d_H(b^a,\,b^n)\right]\geqslant \frac{M}{2}.
\end{equation}

In practice, the hamming distance between similar pairs is not necessarily zero but a
non-negligible very small value $\omega$ due to deficiency in models or characteristics 
of datasets. Suppose it is additive, the expectation above is updated:
\begin{equation}
  \forall (b^a,\,b^n)\in S_N^2,\,\mathbb{E}\left[d_H(b^a,\,b^n)\right]\geqslant \frac{M}{2}+\omega,
\end{equation}
where $\omega\in[0,\,N]$.

Given the $i$-th triple data, with $f^{a_i}$ and $f^{n_i}$ respectively
denoting the corresponding real-valued features, we have
\begin{equation}
  card\left\{u|(f_{u}^{a_i}-0.5)(f_{u}^{n_i}-0.5)<0\right\}\geqslant \left\lceil\frac{M}{2}\right\rceil+\omega.
\end{equation}

As in each dimension, feature value should be $\Delta$ far from 0.5, we have
\begin{equation}
\begin{aligned}
  \forall\,\,u\in\{u|(&f_{u}^{a_i}-0.5)(f_{u}^{n_i}-0.5)<0\} \\
  |f_{u}^{a_i}-f_{u}^{n_i}|=&\left|f_{u}^{a_i}-0.5\right|+\left|0.5-f_{u}^{n_i}\right|>2\Delta.
\end{aligned}
\end{equation}
Setting $\delta$ in Eq. \ref{L_dis} as $4\Delta^2$, we have
\begin{equation}\label{eqn:inequality for alpha_d}
\begin{aligned}
  |f_{u}^{a_i}-f_{u}^{n_i}|^2>\,&4\Delta^2\cdot\left(\left\lceil\frac{\left\lceil \log_2C\right\rceil}{2}\right\rceil
  +\omega\right) \\
  &+\epsilon^2\cdot\left(N-\left\lceil\frac{\left\lceil \log_2C\right\rceil}{2}\right\rceil-\omega\right),    
\end{aligned}
\end{equation}
where $\epsilon\in \left[0,\,0.5\right]$.

Hence, suffice it to set $\alpha_{d}$ as the right part of the inequality above.

\subsection{Implementation}

For implementation, we use the open-source Caffe \cite{jia2014caffe}. All the models are trained by Stochastic
Gradient Descent (SGD) with momentum of 0.9. The mini-batch size is 32 and the weight decay is 0.0005.

The process of training includes two steps.
At the first step, we initialize our network with the pre-trained weights from AlexNet \cite{krizhevsky2012imagenet} and train
our model with the triplet loss for 40 epochs. The margin $\alpha$
is taken as 1.6. The learning rate $\eta$ is set to 0.001.
At the second step, we fine-tune the model with the triplet quantization loss defined in Eq. \ref{L_TQN} for 40 epochs. The learning rate is set to 0.0001. 

In all experiments, $\Delta=0.4$, and hence $\alpha_{sim}=0.16$.
Considering we aim to decrease similar images' distance rather than increase dissimilar images', we set $\beta=8$ and $\gamma=1$ so that the optimization concentrates more on $L_{sim}$.

\section{Experiments}

We verify our model's performance for fast image retrieval on two widely used benchmark data sets of different scales, \textbf{CIFAR-10} \cite{krizhevsky2009learning} and \textbf{In-shop} \cite{liu2016deepfashion}. For convenience of expression, we let ``TQN" denote the proposed method with triplet quantization loss in all the following figures and tables.

\begin{table}
    \centering
    \caption{Parameters for TQN}\label{parameters}
        \begin{tabular}{|c|c|c|c|c|c|c|}
            \hline
            \multirow{2}*{} &\multicolumn{3}{c|}{CIFAR-10 (C=10)} &\multicolumn{3}{c|}{In-shop (C=7,982)} \\
            \cline{2-7}
                &12 bits &24 bits &48 bits &48 bits &96 bits &192 bits \\
            \hline
                $\omega$ &1 &2 &2 &3 &3 &3 \\
            \hline
                $\epsilon$ &0.3 &0.3 &0.3 &0.4 &0.4 &0.4 \\
            \hline
                $\alpha_{d}$ &3.83 &5.46 &7.62 &12.00 &19.68 &35.04 \\
            \hline
        \end{tabular}
\end{table}

\subsection{\textbf{CIFAR-10}}

\textbf{CIFAR-10} contains 10 classes and each class contains 6,000 images. For each class, 5,000 images are randomly picked into training set and rest of them are for test set. For this dataset, we set output as 12, 24 and 48 bits respectively.
The parameter settings are shown in TABLE \ref{parameters}.

Using Mean Average Precision (MAP) as evaluation metric, we compare our models' performance with several state-of-the-arts of supervised hashing methods, including
ITQ, MLH, BRE, KSH, CNNH, DNNH and DSH.

In terms of MAP, our proposed method outperforms all the above hashing methods under comparison as shown in TABLE \ref{res:CIFAR10}. 
Compared with deep hashing methods (CNNH, DNNH and DSH), traditional hashing methods (ITQ, MLH, BRE and KSH) generally have inferior performance.
Furthermore, our proposed method outperforms all these deep hashing methods.
In particular, compared with the best competitor DSH, which is the state-of-the-art deep hashing method leveraging quantization loss, our method increases MAP by 39.02\%, 31.94\%, 27.07\% for 12, 24 and 48 bits, respectively. The experimental results have demonstrated the promise of our proposed triplet quantization loss.
\begin{table}
    \centering
    \caption{Comparison of MAP on CIFAR-10}\label{res:CIFAR10}
        \begin{tabular}{|l|l|l|l|}
            \hline
                Method &12 bits &24 bits &48 bits \\
            \hline
                ITQ &0.165 &0.196 &0.218 \\
                MLH &0.184 &0.199 &0.209 \\
                BRE &0.159 &0.163 &0.172 \\
                KSH &0.295 &0.372 &0.417 \\
            \hline
                CNNH &0.465 &0.521 &0.532 \\
                DNNH &0.555 &0.566 &0.581 \\
                DSH &0.615 &0.651 &0.676 \\
            \hline
                \textbf{Ours} &\textbf{0.855} &\textbf{0.857} &\textbf{0.859} \\
            \hline
        \end{tabular}
\end{table}

\subsection{\textbf{In-shop}}

\textbf{In-shop} contains 52,712 images of 7,982 clothes. 38,494 images are in training set and 14,218 in test set. Images from one clothe have the same label. Given complexity of this dataset, we set output as 48, 96 and 192 bits respectively. 
The parameter settings are shown in TABLE \ref{parameters}. Retrieval accuracy at top 20 is used as the evaluation metric given the characteristic of the data set.

We extract the real-valued features and generate compact binary codes of 48, 96 and 192 bits respectively for convenience of comparison. From TABLE \ref{res:InShop}, where RF denotes real-valued features and BC denotes binary codes, we have the following observations. First, triplet loss is effective in generating highly discriminative real-valued features and leads to the highest retrieval accuracy for all output dimensions. However, its corresponding binary codes have very poor performance with 21.18\%, 18.55\%, and 17.97\% drop in accuracy for 48, 96 and 192 bits, suggesting that the features learned by triplet loss is not suitable for quantization. Secondly, fine-tuning with the triplet quantization loss slightly decreases the disciminability of real-valued descriptors, leading to 5.44\% drop in accuracy for 48 bits, 0.72\% drop for 96 bits, and no drop for 192 bits.
The subsequent quantization of the above real-valued features results in 9.49\%, 6.27\%, and, 1.43\% drop in accuracy, much lower than accuracy drop due to binarizing the features learned with triplet loss, showing the advantage of the triplet quantization loss. Moreover, at 192 bits, the compact binary codes almost have the same retrieval accuracy as the real-valued features.

\begin{table}
    \centering
    \caption{Top 20 retrieval accuracy on In-shop for different output dimensions: 48, 96 and 192 respectively.} \label{res:InShop}
    \begin{tabular}{|c|c|c|c|}
        \hline
            &48 &96 &192 \\
        \hline
            RF with triplet &0.680 &0.690 &0.701 \\
        \hline
            BC with triplet &0.536 (-21.18\%) &0.562 (-18.55\%) &0.575 (-17.97\%) \\
        \hline
            RF with TQN &0.643 (-5.44\%) &0.685 (-0.72\%) &0.701 (-0.00\%) \\
        \hline
            BC with TQN &0.582 (-14.41\%) &0.642 (-6.96\%) &0.691 (-1.43\%) \\
        \hline
    \end{tabular}
\end{table}

\begin{figure}
  \centering
  \includegraphics[width=2.7in]{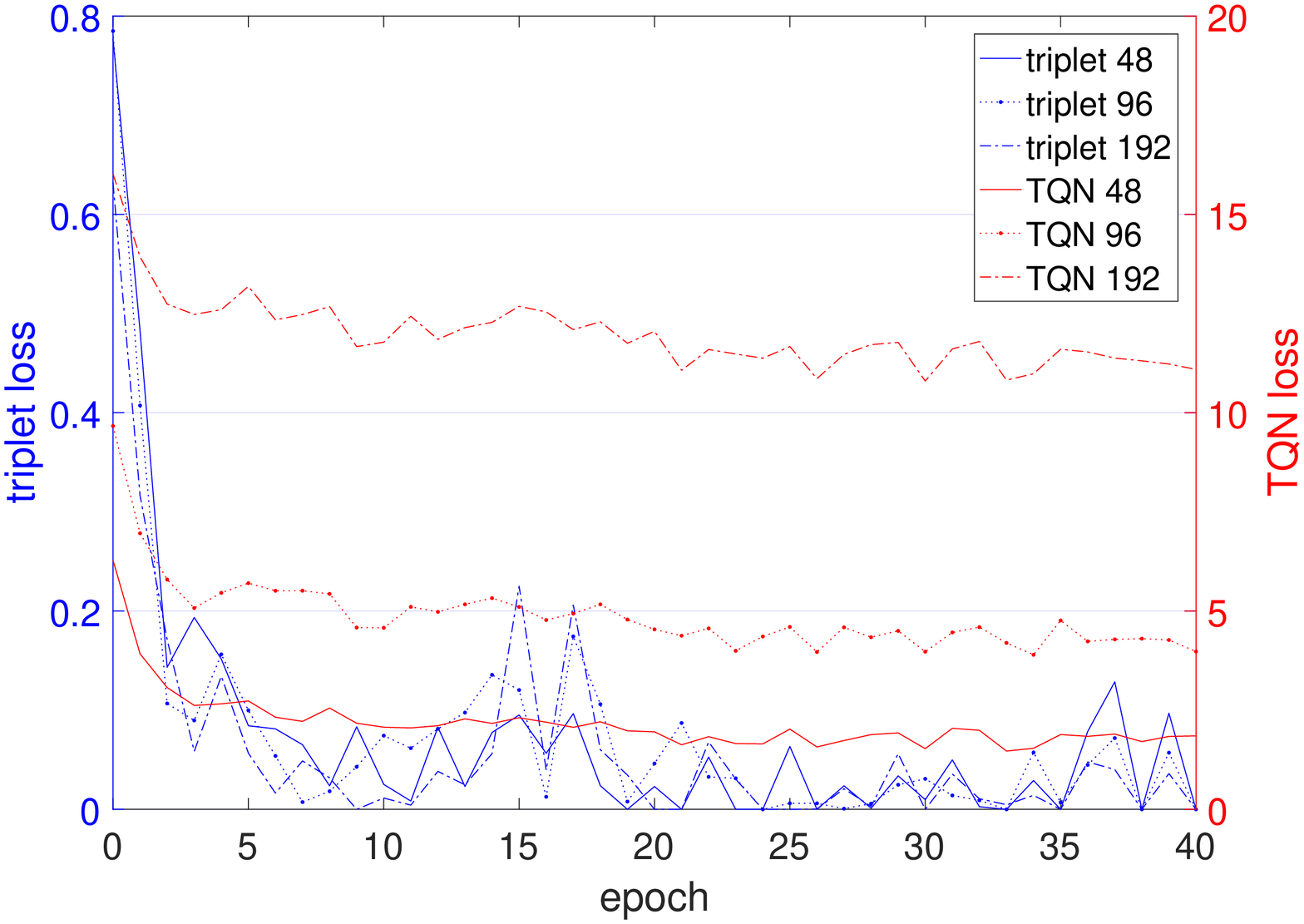}
  \caption{Change of triplet loss and triplet quantization loss for different output dimensions during the first step.
  }\label{lossChange}
\end{figure}

\begin{figure}
    \centering
    \includegraphics[width=2.7in]{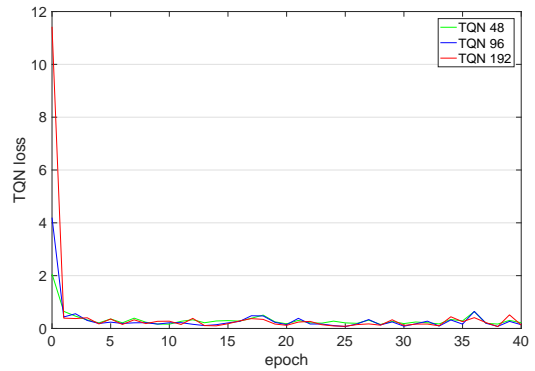}
    \caption{Change of triplet quantization loss for different output dimensions during the second step.}
    \label{TQNlossChange}
\end{figure}

Fig. \ref{lossChange} and Fig. \ref{TQNlossChange} show how losses change during the two steps. At the first step where the triplet loss is used, the triplet loss is effectively lowered, but the triplet quantization loss maintains at a high level (Fig. \ref{lossChange}). At the fine-tuning step, the triplet quantization loss decreased very fast, as shown in Fig. \ref{TQNlossChange}. Considering retrieval accuracy of binary codes has been significantly improved after fine-tuning, it is reasonable to conclude that the triplet quantization loss successfully optimizes loss of information during quantization. 
Fig. \ref{TQNlossChange} shows that the triplet quantizaion loss converges rapidly and keeps at almost the same low level for different bits, suggesting that the parameter settings are reasonable so that the optimization is achieved in very few iterations.

\section{Conclusion}

In this paper, we have proposed a supervised deep hashing method for generating discriminative binary codes.
A loss function, triplet quantization loss, is designed to keep the distance of similar images smaller than that of dissimilar images after quantization. We also propose a reasonable way of setting parameters, with which the proposed loss converges rapidly. Experiments on two widely used benchmarks have shown that our proposed method has superior retrieval results over state-of-the-art deep hashing methods, especially compared with the ones utilizing quantization loss.

\section{Acknowledgements}
The work is partially supported by the High Technology Research and Development Program of China 2015AA015801, NSFC 61521062, STCSM 12DZ2272600.

\bibliographystyle{IEEEtran}

\bibliography{IEEEabrv,IEEEexample}

\end{document}